\PassOptionsToPackage{noend}{algorithmic}

\documentclass{article}
\pdfoutput=1

\usepackage{etoolbox}
\newtoggle{anonymize}
\newtoggle{arxiv}

\settoggle{arxiv}{true}
\settoggle{anonymize}{false}

\iftoggle{arxiv}{%
	\usepackage[preprint]{neurips_2019}
}{
	\iftoggle{anonymize}{%
		\usepackage{neurips_2019}
	}{

		\usepackage[final]{neurips_2019}

	}
}

\usepackage{amsmath,amsfonts,bm}

\def\eqref#1{equation~\ref{#1}}

\def\1{\bm{1}}

\def\vf{{\bm{f}}}
\def\vg{{\bm{g}}}

\def\vp{{\bm{p}}}

\def\vw{{\bm{w}}}

\def\evf{{f}}
\def\evg{{g}}

\DeclareMathAlphabet{\mathsfit}{\encodingdefault}{\sfdefault}{m}{sl}
\SetMathAlphabet{\mathsfit}{bold}{\encodingdefault}{\sfdefault}{bx}{n}

\newcommand{\E}{\mathbb{E}}

\usepackage{graphicx}
\usepackage{graphbox}

\usepackage[utf8]{inputenc} 
\usepackage[T1]{fontenc}    
\usepackage{hyperref}       
\usepackage{url}            
\usepackage{booktabs}       
\usepackage{amsfonts}       
\usepackage{nicefrac}       
\usepackage{microtype}      

\usepackage{subcaption}
\usepackage{cleveref} 
\usepackage{wrapfig}
\usepackage{xr}

\usepackage{algorithm}
\usepackage{algorithmic}

\usepackage{siunitx}
\newcommand{\Defined}{\overset{\text{def}}{=}}
\newcommand{\EqRef}[1]{(\ref{eq:#1})}
\newcommand{\FigRef}[1]{Figure~\ref{fig:#1}}

\title{Toward Runtime-Throttleable Neural Networks}

\author{%
  Jesse Hostetler\\
  Center for Vision Technology\\
  SRI International\\
  Princeton, NJ 08540 \\
  \texttt{jesse.hostetler@sri.com} \\
}

\begin{document}

\maketitle

\begin{abstract}
As deep neural network (NN) methods have matured, there has been increasing interest in deploying NN solutions to ``edge computing'' platforms such as mobile phones or embedded controllers. These platforms are often resource-constrained, especially in energy storage and power, but state-of-the-art NN architectures are designed with little regard for resource use. Existing techniques for reducing the resource footprint of NN models produce static models that occupy a single point in the trade-space between performance and resource use. This paper presents an approach to creating \emph{runtime-throttleable} NNs that can adaptively balance performance and resource use in response to a control signal. Throttleable networks allow intelligent resource management, for example by allocating fewer resources in ``easy'' conditions or when battery power is low. We describe a generic formulation of throttling via block-level \emph{gating}, apply it to create throttleable versions of several standard CNN architectures, and demonstrate that our approach allows smooth performance throttling over a wide range of operating points in image classification and object detection tasks, with only a small loss in peak accuracy.
\end{abstract}

\section{Introduction}
State-of-the-art deep neural network (NN) architectures for computer vision problems are extremely resource-hungry. This poses a problem for NN-based machine learning systems deployed on ``edge'' devices such as mobile phones or embedded controllers that are constrained in terms of computation, energy storage, and heat dissipation. The best-performing architectures are too big to run on such platforms. In some applications, the heavy computation can be off-loaded to a remote server, but for applications like remote sensing, industrial controls, or autonomous robots, it is not acceptable for the system to depend on a potentially-unreliable network connection.

Recognizing these difficulties, much recent work has focused on reducing the resource requirements of NN models, by tweaking the network architecture \citep[e.g.][]{howard2017mobilenets}, substituting low-rank approximations of the weight tensors \citep[e.g.][]{lebedev2015speeding}, or quantizing the weights to a lower precision \citep[e.g.][]{rastegari2016xnor}. While each of these approaches has succeeded to some degree, they share the limitation of producing a \emph{single} simplified model that occupies one point in the trade-space between performance and resource use. In reality, the ``best'' model for an application is determined by conditions that change over time. For example, a surveillance system may have lower accuracy in low-light conditions, and it would be useful to ``throttle up'' its performance while the challenging conditions persist. Likewise, a battery-powered sensor could ``throttle down'' its performance to extend its battery life, even at the expense of failing to detect some interesting events.

We present an approach to creating \emph{throttleable} neural networks whose performance can be varied at run-time to fit the situation. We do this by partitioning the network into a set of disjoint components and training the network in such a way that individual components can be ``turned off'' with minimal loss of accuracy. A separate \emph{gating module} decides which components to turn off to obtain the best performance for a given level of ``computational effort.'' Our method is largely model-agnostic, as we demonstrate by applying it to several different convolutional NN (CNN) architectures. Importantly, due to our focus on computational ``blocks'', our method is also amenable to acceleration in hardware because it preserves most opportunities for vectorized computation that are present in the original architectures. Our two-stage training approach allows the ``data path'' of the network to be trained once, while the gating module -- which is computationally much simpler -- can be trained and re-trained separately. This makes the overall throttleable NN highly adaptable, and the gating module could even potentially be trained on the edge device to accommodate a changing environment.

In the remainder of this paper, we describe our approach to throttleable NNs, show how to create throttleable versions of different CNN architectures, and discuss several schemes for gating. We then evaluate different TNN architectures on image classification and object detection tasks.

\begin{figure}[t]
\captionsetup[subfigure]{position=b}
\setbox9=\hbox{\includegraphics[width=0.4\columnwidth]{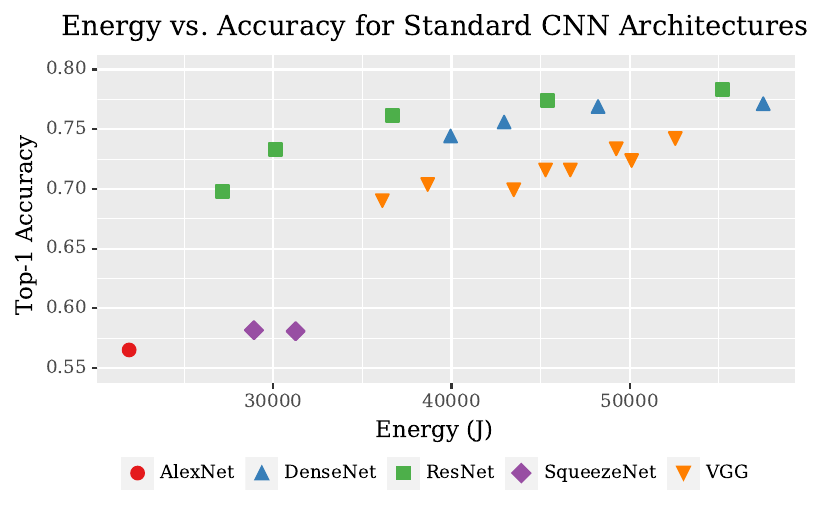}}
\centering
\subcaptionbox{\label{fig:energy}}{
	\includegraphics[width=0.4\columnwidth]{fig/profile_imagenet_0.pdf}
	\vspace{-0.2cm}
}
\hfill
\subcaptionbox{\label{fig:width-depth}}{
	\raisebox{\dimexpr\ht9-\height}{\includegraphics[width=0.24\columnwidth]{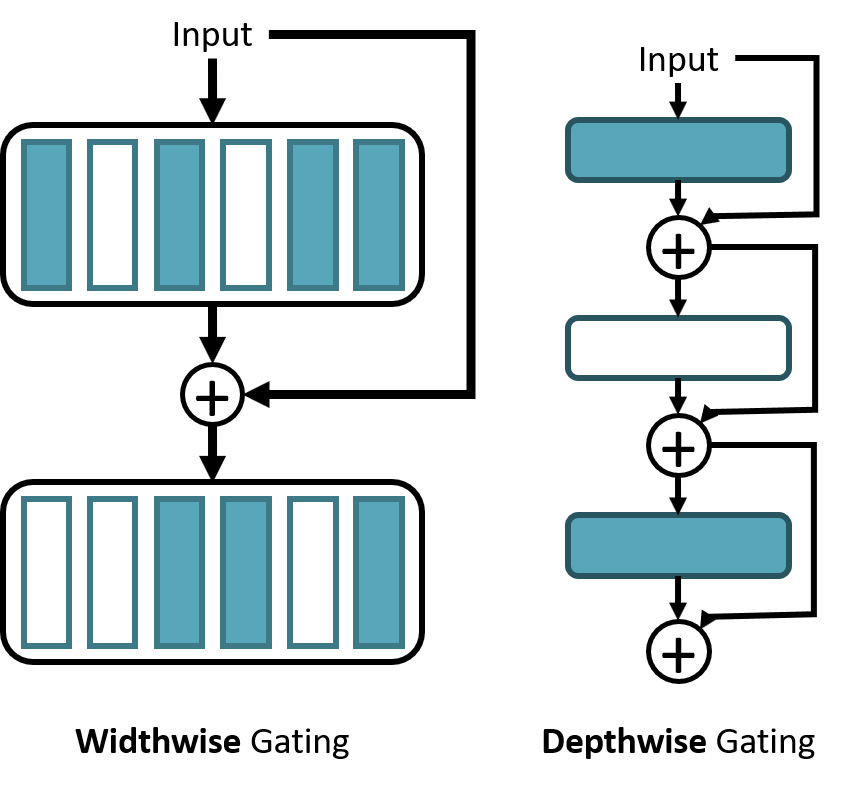}}
	\hfill
	\raisebox{\dimexpr\ht9-\height}{\includegraphics[width=0.3\columnwidth]{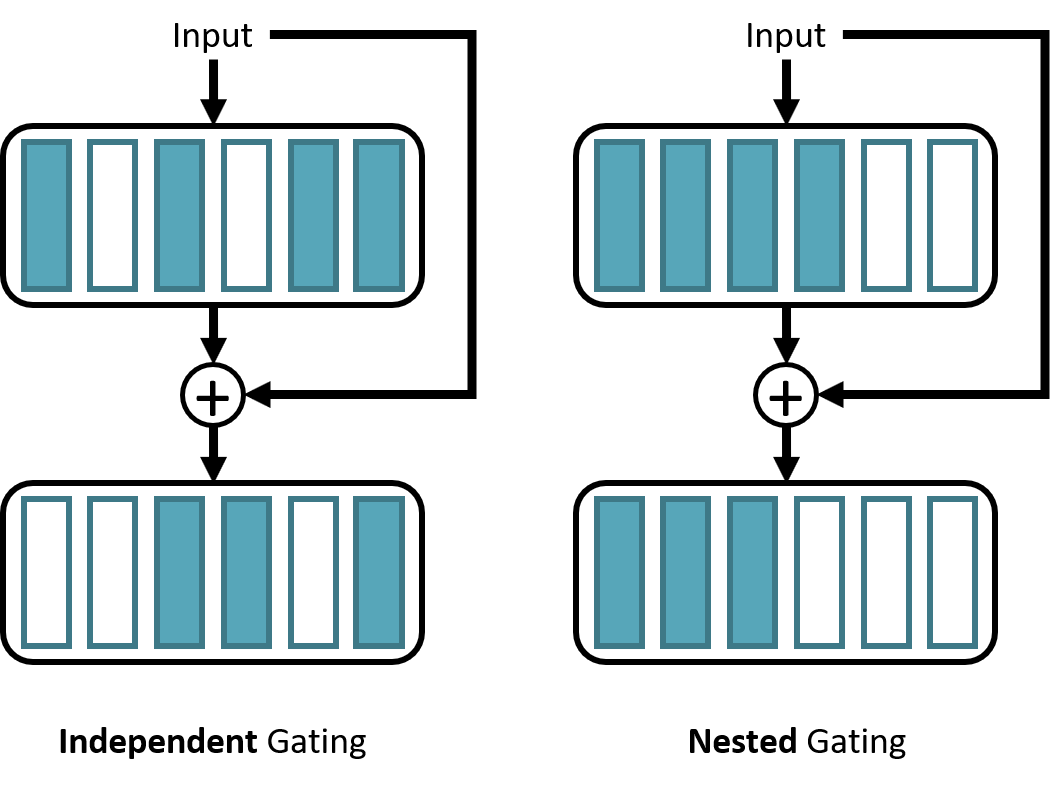}}
	\vspace{-0.2cm}
}
\begin{NoHyper}
\caption{(\subref{fig:energy}) Neural network architectures are a compromise between resource consumption and performance. This figure shows energy consumption vs. top-1 accuracy for the ImageNet validation set on an Nvidia GTX 1080 Ti GPU (mean of 2 runs on different hardware). The goal of Throttleable Neural Networks is to train a single model for which this trade-off can be varied at runtime. (\subref{fig:width-depth}) This paper considers performance throttling via selective \emph{gating}. This figure shows different ways of organizing gated components -- \emph{widthwise} vs. \emph{depthwise} gating, and \emph{independent} vs. \emph{nested} ordering.}
\end{NoHyper}
\end{figure}

\section{Throttleable Neural Networks}
\label{sec:throttleable}

A neural network is a parameterized function $h_{\theta}(x)$ mapping an input $x \in X$ to an output $y \in Y$. We define a \emph{throttleable neural network} (TNN) as a function of two variables, $h_{\theta}(x, u)$, where $u \in [0,1]$ is a control parameter that indicates how much ``computational effort'' the network should exert. We emphasize that $u$ is an additional input to the network; after training is complete, the parameters $\theta$ are fixed but $u$ can change. In our framework, the loss function of a TNN has two components,
\begin{equation} \label{eq:tnn-objective}
J(x, u, y, \hat{y}) = L(y, \hat{y}) + \lambda C(x, u).
\end{equation}
The ``task loss'' component, $L$, is a task-specific performance measure, e.g., cross-entropy loss for classification. The ``complexity loss,'' $C$, measures the resources used -- energy, CPU time, etc. -- when the network processes example $x$ at ``effort level'' $u$, and $\lambda$ controls the balance of the two losses. A simple example of a TNN would be an architecture that employs Dropout \citep{srivastava2014dropout} at test-time. In this case, the dropout probability $u$ would be the control parameter. We develop more sophisticated schemes in Section~\ref{sec:architectures}.

\subsection{Related Work}
The only work we are aware of that addresses runtime-controllable throttling of NNs is that of \citet{odena2017changing}. That work proposes a model in which each layer has multiple data paths, and a ``Composer'' module chooses which path to take in each layer. The Composer takes a control parameter as input and its loss function penalizes complexity weighted by the control signal. We describe a broader TNN framework that subsumes the model of \citet{odena2017changing}. There is a much larger body of work on static approaches to reducing NN resource requirements.

\paragraph{Model Compression} Works such as MobileNets \citep{howard2017mobilenets} have demonstrated how small tweaks to an architecture can greatly reduce resource requirements with minimal loss of accuracy. One family of approaches that performs such a transformation automatically is quantization, which uses reduced-precision for weights and/or activations to reduce memory usage and, in some cases, enable cheaper computation in hardware. Examples include Binarized NNs \citep{courbariaux2015binaryconnect, hubara2016binarized}, XNOR-Nets \citep{rastegari2016xnor}, and (Generalized) Ternary Connect \citep{lin2016neural,parajuli2018generalized}. Another approach is to remove computations without affecting the result, such as by pruning near-$0$ weights \citep[e.g.][]{han2015learning} or using low-rank weight tensor approximations \citep{lebedev2015speeding}. Other schemes are based on structuring the computations in the network to use available hardware elements most effectively \citep[e.g.][]{lane2016deepx}.

\paragraph{Conditional Computation} Conditional computation or ``gating'' is based on turning off parts of the network. This can be viewed as ``block-wise'' dropout \citep{srivastava2014dropout} applied at runtime. \citet{bengio2013estimating} and \citet{bengio2015conditional} consider stochastic gating with Bernoulli random variables. The sparsely-gated mixture-of-experts model \citep{shazeer2017outrageously} learns to ``rank'' NN modules and selects only the top $k$ modules for computation. Many conditional computation schemes use ResNet modules \citep{he2016deep} as building blocks, leveraging the observation that ResNets behave as an ensemble of shallower networks
\citep{veit2016residual}. Skipnet \citep{wang2017skipnet} and Blockdrop \citep{wu2018blockdrop} are very similar approaches that learn to bypass ResNet blocks based on the input. Similar ideas also appear in recent work on neural architecture search \citep[e.g.][]{pham2018efficient}. A notable subclass of conditional computation approaches is based early stopping once some threshold of ``confidence'' is achieved. Examples of this approach include Adaptive Computation Time (ACT/SACT) \citep{figurnov2017spatially}, BranchyNet \citep{teerapittayanon2016branchynet}, amd Dynamic Time Recurrent Visual Attention (DT-RAM) \citep{li2017dynamic}.

\section{Throttling via Gating}
\label{sec:architectures}

The design space of TNNs is very large. In this paper, we develop throttleable versions of common CNN building blocks. If throttling is to reduce resource use, the function that determines how to throttle the network should be much cheaper to compute than the network that it controls. It is not practical, for example, to control throttling at the level of individual neurons. Our approach is based on structuring the network into ``blocks'' of neurons and applying gating at the block level.

\subsection{Modular Gated Networks}
We consider a family of TNN architectures that we call \emph{modular gated networks}. A \emph{gated module} has the functional form
\begin{equation*}
y = a(\vg_{\psi}(x, u) \odot \vf_{\theta}(x)),
\end{equation*}
where $\vf_{\theta}(x) = (f_1, \ldots, f_n)$ is a vector of \emph{components} with parameters $\theta$, $\vg_{\psi}(x, u) : X \times [0,1] \mapsto \{0,1\}^n$ is the \emph{gating function} with parameters $\psi$, $\odot$ denotes element-wise multiplication, and $a$ is the \emph{aggregation function} that maps $\vg_{\psi}(x, u) \odot \vf_{\theta}(x)$ to the appropriate output space. The elements of $\vf$ can be arbitrary NN modules, but we assume they have the same input space and that their outputs can be aggregated appropriately, such as by concatenating or summing them. We will omit the parameter subscripts when they are not relevant. Our exposition focuses on a single gated module for simplicity, but note that in practice we compose multiple gated modules to create a typical multi-layer NN.

As usual when applying dropout, we found it necessary to normalize the activations so that the output magnitude is similar for all dropout rates. Thus in practice, we implement gated modules of the form
\begin{equation} \label{eq:gated-module}
y = a(\bar{\vg}_{\psi}(x, u) \odot \vf_{\theta}(x)),
\end{equation}
where $\bar{\vg}$ is the normalized gating function,
\begin{equation}
\bar{\vg}(x, u) \Defined \frac{\vg(x, u)}{||\vg(x, u)||_1}.
\end{equation}

When $\evg_i = 0$, the component $\evf_i$ is effectively disabled. When training on a GPU, we implement the mathematical form \EqRef{gated-module} directly to take advantage of vectorized computations. In a deployed system, we would skip computing $\evf_i$ when $\evg_i = 0$ to realize power savings.


\subsection{Defining the Components}
The components of $\vf$ could be anything from individual neurons to entire networks. We focus on an intermediate level of granularity. The size of neural networks can be measured along two dimensions: the ``width'' or number of features per layer, and the ``depth'' or number of layers from input to output. Decompositions into components can be defined along both of these dimensions (Figure~\ref{fig:width-depth}).

\paragraph{Width-wise gating} entails disabling some of the neurons in a single layer. To reduce the complexity of the gating function, we consider \emph{block} gating schemes in which neurons are partitioned into disjoint sets and gated as a group. In convolution layers, the blocks are sets of convolutional filters; in fully-connected layers, the blocks are sets of neurons. Examples of width-wise gating include the mixture-of-experts layer \citep{shazeer2017outrageously} and the Composer model \citep{odena2017changing}.

\paragraph{Depth-wise gating} entails bypassing entire layers. It is applicable only to networks that have skip-connections, since otherwise disabling a layer would cause all subsequent layers to receive no information about the input. This is the more common form of gating in recent literature \citep[e.g.][]{huang2016deep,figurnov2017spatially,wang2017skipnet,wu2018blockdrop}, where it is usually applied in a ResNet-style architecture \citep{he2016deep}.

\begin{table}
\caption{Summary of the throttleable CNN architectures used in our experiments. (*) Throttleable DenseNets have qualities of both widthwise and depthwise gating.}
\label{tab:architectures}
\centering
\begin{tabular}{|l|c|c|c|}
\hline
Architecture & Axis & Components & Aggregation \\
\hline
VGG & Width & Conv. Filters & Concat \\
ResNeXt-W & Width & Conv. Filters & Sum \\
ResNet-D & Depth & Layers & Sum \\
DenseNet & Width* & Layers & Concat \\
\hline
\end{tabular}
\end{table}

\subsection{Throttleable CNN Architectures}
To examine the generality of our TNN concept, we created throttleable versions of several popular CNN architectures, summarized in Table~\ref{tab:architectures}.

\paragraph{VGG} The VGG architecture \citep{simonyan2014very} is a typical example of a ``single-path'' CNN. We apply width-wise gating to groups of convolutional filters in each layer and combine the group outputs by concatenating them. Because VGG lacks skip-connections, we enforce that at least one group must be active in each layer, to avoid making the output zero.

\paragraph{ResNeXt-W} ResNeXt \citep{xie2017aggregated} is a modification of ResNet \citep{he2016deep} that structures each ResNet layer into groups of convolutional filters that are combined by summing. We created a widthwise-gated version of ResNeXt (``ResNeXt-W'') by treating each filter group as a gated component. We suspected that this architecture is particularly well-suited for width-wise gating, since the summing operation is ``smoother'' than concatenation.

\paragraph{ResNet-D} We also experimented with a depthwise-gated version of standard ResNet (``ResNet-D''), similar to Blockdrop/Skipnet \citep{wang2017skipnet,wu2018blockdrop}. In this architecture, the gated components are entire ResNet blocks that are skipped when gated off.

\paragraph{DenseNet} In the DenseNet architecture \citep{huang2017densely}, each dense block contains multiple narrow layers that are combined via concatenation. These narrow layers make natural units for gating. We view this architecture as primarily widthwise-gated since the components are concatenated ``horizontally,'' but it also has qualities of depthwise gating due to the skip connections.

\subsection{Order of Gating}
In all other work on gated networks that we are aware of, the components of each gated module are viewed as independent of one another, with few constraints on their pattern of activation. This \emph{independent gating} scheme makes sense when the goal is for each component to model different features of the data, such as in a mixture-of-experts architecture \citep{shazeer2017outrageously}, and there is some evidence that independent contextual gating induces this type of specialization \citep{wu2018blockdrop}. For our goal of throttling over a range of set points, however, this specialization is not necessary and may be counterproductive, since it can be expected to produce some redundancy in the representation. We propose a different method that we call \emph{nested gating}. In the nested scheme, the gating function $\vg$ is constrained such that $g_i > 0 \Rightarrow g_j > 0 \; \forall j < i$ (\FigRef{width-depth}). Empirically, we observe that the nested scheme gives superior throttling performance given the same architecture (Section~\ref{sec:experiments}).

\section{Training Throttleable Networks}

The goal of training a throttleable network is to create a model that varies its complexity in response to the control parameter $u$. The natural measure of complexity is the number of active components, possibly weighted by some measure of resource consumption for each component,
\begin{equation}
c(\vg) = ||\vw||_1^{-1} \sum_i w_i \mathbf{1}(g_i \neq 0).
\end{equation}

The \emph{gate control strategy} embodied in $\vg_{\psi}(x, u)$ modulates the resource utilization of the TNN. Our experiments examine both static and learned gating functions. In the static approaches, the control parameter $u$ determines the number of gated blocks that should be used, and the choice of which blocks to turn on is made according to a fixed rule. Empirically, we find that a straightforward application of the \emph{nested} gating order works very well.


We also consider \emph{learning} the gating function. We enforce the constraint that the actual complexity $c(\vg)$ should not exceed the target complexity $u$ by optimizing the combined loss function $J(x, u, y, \hat{y}) = L(y, \hat{y}) + \lambda C(x, u)$. We experimented with variants of $C$ of the two functional forms
\begin{align}
C^p_{\text{hinge}}(x, u) & \Defined \text{max}(0, c(\vg(x,u)) - u)^p \label{eq:hinge} \\
C^p_{\text{dist}}(x, u) & \Defined |c(\vg(x,u)) - u)|^p, \label{eq:dist}
\end{align}
for $p \in \{1,2\}$. In a sense, the ``hinge'' penalty \EqRef{hinge} is the ``correct'' objective, since there is no reason to force the model to use more resources unless it improves the accuracy. In practice, we found that the ``distance'' penalty resulted in somewhat higher accuracy for the same resource use.

Learning the gate controller is complicated by the ``rich get richer'' interaction between $\vg$ and $\vf$, in which only the subset of $\vf$ selected by $\vg$ receives training, which improves its performance and reinforces the tendency of $\vg$ to select it. 
To address this, we adopt a two-phase training strategy similar to \citet{figurnov2017spatially}.
In the first phase, we train the ``data path'' with random gating to optimize only $L$ while being ``compatible'' with gating. In the second phase, we train the gating controller to optimize the full objective $J$ while keeping the data path fixed.



\subsection{Training the Data Path}
\begin{figure}
\begin{minipage}{0.42\columnwidth}
\begin{algorithm}[H]
\begin{algorithmic}
	\STATE \COMMENT{Train the data path}
	\FOR{\# epochs in Phase 1}
		\FORALL{$(x, y)$ in training data}
			\STATE Let $k \sim \text{DiscreteUniform}[0, n]$
			\STATE Let $\vg = 1^k 0^{n-k}$ 
			\IF{using \emph{independent gating}}
				\STATE $\vg \leftarrow \text{RandomPermutation}(\vg)$
			\ENDIF
			\STATE Let $\hat{y} = a(\bar{\vg} \odot \vf_{\theta}(x))$
			\STATE $\theta \leftarrow \theta - \alpha \nabla L(y, \hat{y})$
		\ENDFOR
	\ENDFOR
	\STATE \COMMENT{Train the gate controller}
	\FOR{\# epochs in Phase 2}
		\FORALL{$(x, y)$ in training data}
			\STATE Let $u \sim \text{Uniform}[0, 1]$
			\STATE Let $\hat{y} = a(\bar{\vg}_{\psi}(x, u) \odot \vf_{\theta}(x))$
			\STATE $\psi \leftarrow \psi - \alpha \nabla J(x, u, y, \hat{y})$
	    \ENDFOR
	\ENDFOR
\end{algorithmic}
\caption{Two-Phase TNN Training}
\label{alg:data}
\end{algorithm}
\end{minipage}
\hfill
\begin{minipage}{0.54\columnwidth}
\centering
\includegraphics[width=0.9\columnwidth]{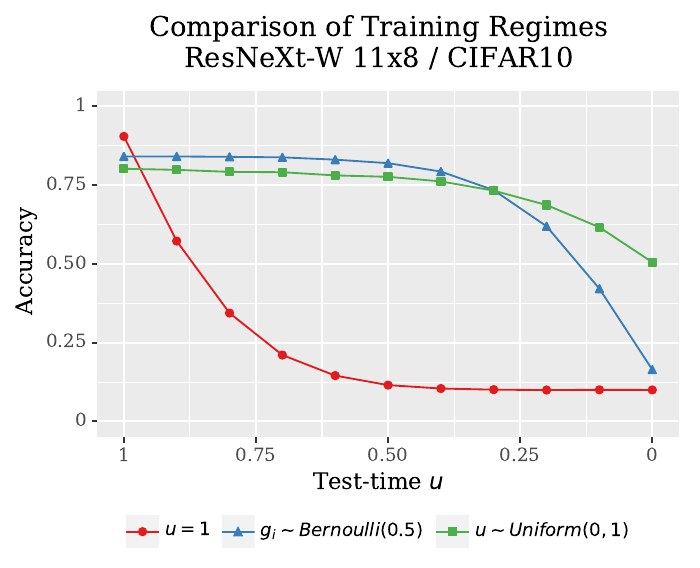}
\caption{Accuracy with runtime throttling for different gating regimes at training time.}
\label{fig:dropout}
\end{minipage}
\end{figure}

During Phase~1 of training, we train the feature representations of the TNN to be robust to varying amounts of gating. The choice of how $u$ is sampled during training is important for obtaining the desired performance profile. From an empirical risk minimization perspective, we can interpret the training-time distribution of $u$ as a prior distribution on the values of $u$ that we expect at test-time. Ordinary training without gating can be viewed as one extreme, where we always set $u = 1$. In Figure~\ref{fig:dropout}, we compare these three schemes on the CIFAR10 dataset \citep{krizhevsky2009learning} using a simple ResNeXt model. We can see that there is a trade-off between peak performance and ``average-case'' performance as quantified by the area under the curve. Thus, we should choose the training-time distribution of $u$ to match the anticipated test-time usage profile.

In our experiments, we employ a training scheme designed to maximize the useful range of $u$. For each training example, we draw $u \sim \text{Uniform}[0,1]$. Then, for each gated module, we select $k$ blocks to be gated on, where $k = \text{min}(n, \lfloor u \cdot (n+1) \rfloor)$ and $n$ is the number of blocks in the module. For \textsc{Nested} gating strategies (\FigRef{width-depth}), we set $g_1, \ldots, g_k$ to $1$ and $g_{k+1}, \ldots, g_n$ to $0$, while for \textsc{Independent} gating strategies we choose $k$ indices at random without replacement.


\subsection{Training the Gating Module}

When learning $\vg$, we proceed to Phase~2 of training, where we hold the data path parameters $\theta$ fixed and optimize the gate controller parameters $\psi$. As in Phase~1, we draw the target utilization $u$ from a uniform distribution. We model the components of the gating function as Bernoulli random variables,
\begin{equation} \label{eq:gbern}
g_i(x, u; \psi) \sim \text{Bernoulli}(p_i(x, u; \psi)),
\end{equation}
and our task is to learn the function $\vp_{\psi}$ giving the activation probabilities of each component. Since $C$ is discontinuous, we need to employ a gradient estimator for training. We evaluated two existing methods of training networks with stochastic discrete neurons for this purpose.


\paragraph{Score function estimator} The most common approach \citep{bengio2013estimating, wang2017skipnet, wu2018blockdrop} is to treat $\vg$ as the output of a stochastic policy and train it with a policy gradient method such as the score function (\textsc{REINFORCE}) estimator,
\begin{equation*}
\nabla_{\psi} \E[J] = \E[J \cdot \nabla_{\psi} \log \text{Pr}(\vg_{\psi}(x, u))],
\end{equation*}
where $\text{Pr}(\vg_{\psi}(x, u))$ is the density of the random variable $\vg$. Since each $g_i$ is an independent Bernoulli random variable \EqRef{gbern}, the log probability is given by $\log \text{Pr}(\vg) = \sum_i \log [g_i p_i + (1 - g_i) (1 - p_i)]$.

\paragraph{Continuous relaxations} Relaxation approaches soften the discrete gate vector into a continuous vector of ``activation strengths.'' In particular, we use Concrete random variables \citep{maddison2017concrete} to stand in for discrete gating during training. Concrete distributions have a temperature parameter $t$ where the limit $t \rightarrow 0$ recovers a corresponding discrete distribution. The Bernoulli distribution is replaced by the binary Concrete distribution,
\begin{equation*}
\evg_i \sim \sigma((L + \log \alpha_i)^{-t}),
\end{equation*}
where $L \sim \text{Logistic}(0,1)$ and $\alpha_i = p_i/(1-p_i)$. We set $t > 0$ during training to make the network differentiable, and use $t = 0$ during testing to recover the desired hard-gated network.

\begin{figure*}[t!]
\centering
\includegraphics[width=\textwidth]{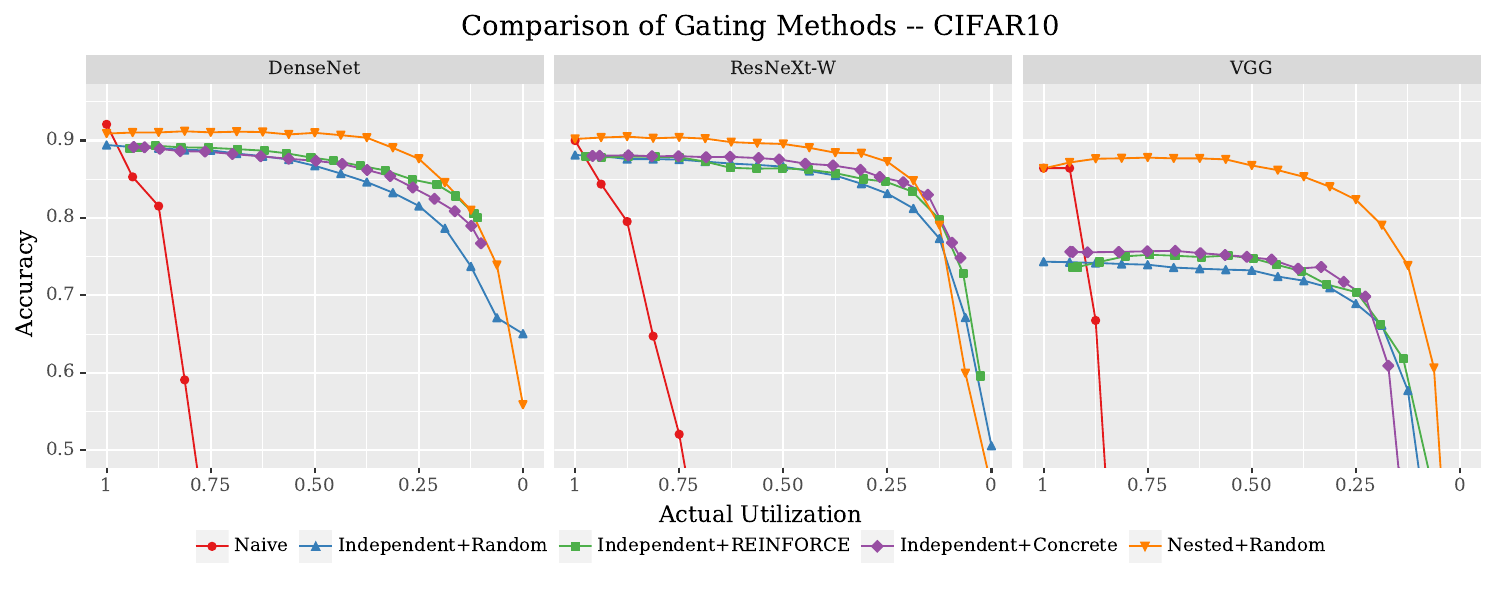}
\label{fig:classification}
\vspace{-0.5cm}
\caption{Comparison of classification accuracy with different gate control methods for three standard CNN architectures on the CIFAR10 dataset.}
\label{fig:cifar10-blind}
\end{figure*}

\section{Experiments} \label{sec:experiments}
Our experiments compare different approaches to creating TNNs using gating in image classification and object detection tasks. Our goal is not to achieve state-of-the-art performance in any task, but rather to make a fair comparison between gating strategies using representative NN models and tasks. Additional details of our methods are available in the supplementary material. 


\subsection{Image Classification: CIFAR10}
The CIFAR10 dataset \citep{krizhevsky2009learning} is a standard image classification benchmark consisting of 32x32 pixel color images of 10 categories of object. We used the standard 50k image training set and 10k image test set, with no data augmentation. The CNN architectures were as follows. \textbf{DenseNet:} DenseNet-BC with 3 dense blocks having 16 components each with a growth rate of 12. \textbf{ResNeXt:} The ResNeXt architecture for CIFAR as described by \citet{xie2017aggregated} with 16 gated components in each of the 3 stages. \textbf{VGG:} The VGG-D architecture truncated to the first 3 convolution stages followed by a 4096 unit fully-connected layer; all three convolution stages and the fully-connected layer were partitioned into 16 gated components. The ``Independent+Learner'' methods use a ``blind'' control network (FC $\rightarrow$ ReLU $\rightarrow$ FC) that maps the control input $u$ to gate vectors $\vg$ for each gated module. We show results for the $C^2_{\text{dist}}$ complexity penalty (\ref{eq:dist}) and $\lambda = 10$.

\iftoggle{arxiv}{%
	\newcommand\UsageFigRef{Figure~\ref{fig:usage-cifar10-densenet}}
}{
	\newcommand\UsageFigRef{Supplement Figure~\ref{supp-fig:usage-cifar10-densenet}}
}

\paragraph{Results} The most noticeable result is that \emph{nested gating} substantially outperformed all variations on the independent method for all 3 architectures (\FigRef{cifar10-blind}). The difference is especially pronounced for VGG; we attribute this to VGG learning more ``entangled'' representations than architectures with skip connections, which could make it more sensitive to exactly which components are gated off. For independent gating, the learned gating controllers were consistently better than random gating for both REINFORCE and Concrete training methods. The learned controllers achieve better performance by allocating computation non-uniformly across the different stages of the network (\UsageFigRef). Note that the learned gating functions do not cover the entire range of possible utilization $[0,1]$. The useful range of $u$ was larger for larger $\lambda$ and for complexity penalties with $p = 1$, but these also resulted in lower accuracy overall.

\subsection{Image Classification: ImageNet} Our second set of experiments examines image classification on the larger-scale ImageNet dataset \citep{russakovsky2015imagenet} using the DenseNet-169, ResNeXt-50, and ResNet-50 architectures. For ImageNet, we used pre-trained weights to initialize the data path, then fine-tuned the weights with gating. We used the DenseNet-169 and ResNet-50 models from the \texttt{torchvision} package of PyTorch, and for ResNeXt-50 we converted the original Torch model \citep{xie2017aggregated} to PyTorch using a conversion utility \citep{clcarwin2017convert}.

\begin{figure}[t]
\captionsetup[subfigure]{position=b}
\setbox9=\hbox{\includegraphics[width=0.4\columnwidth]{fig/profile_imagenet_0.pdf}}
\centering
\subcaptionbox{\label{fig:imagenet}}{
	\includegraphics[width=0.48\columnwidth]{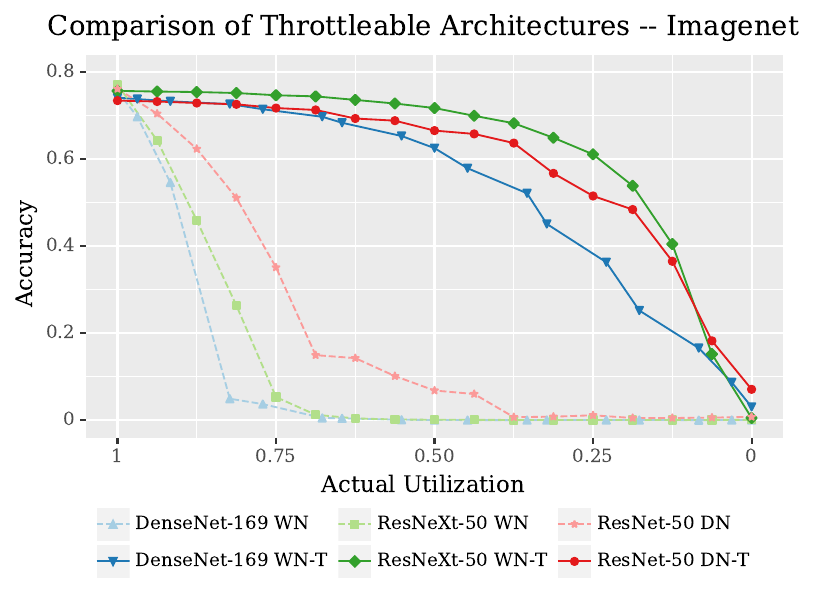}
	\vspace{-0.4cm}
}
\hfill
\subcaptionbox{\label{fig:voc2007}}{
	\includegraphics[width=0.48\columnwidth]{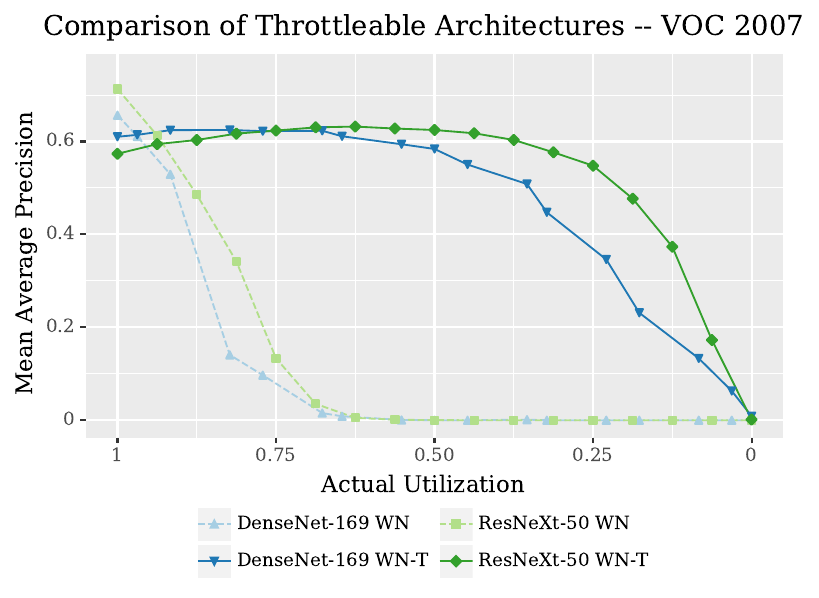}
	\vspace{-0.4cm}
}
\begin{NoHyper}
\caption{Comparison of throttleable architectures on larger-scale vision tasks. In the legend, ``WN'' (``DN'') means \emph{widthwise} (\emph{depthwise}) \emph{nested} gating. The ``-T'' suffix indicates finetuning with \emph{throttling} (Phase~1 of Algorithm~\ref{alg:data}). (\subref{fig:imagenet}) Classification performance on ImageNet. (\subref{fig:voc2007}) Object detection on VOC2007, using the Faster-RCNN architecture with a TNN as the ``backbone''.}
\end{NoHyper}
\end{figure}

In these experiments, we consider widthwise nested gating (``WN'' in Figure~\ref{fig:imagenet}) and depthwise nested gating (``DN''). In the DN scheme, we repeatedly iterate through the stages of the ResNet network from output to input and turn on one additional layer in each stage, unless the proportion of active layers in that stage exceeds $u$, and stop when the total utilization exceeds $u$. The ``-T'' suffix in the figures indicates that fine-tuning with gating was applied.

\paragraph{Results} The throttleable models reached a peak accuracy within about $2-3\%$ of the corresponding pre-trained model, and all were smoothly throttleable through the full range of utilization whereas the pre-trained models degrade rapidly with increased throttling. The ResNeXt model was best in terms of both peak accuracy and area-under-curve. Unlike in CIFAR10, the DenseNet model was substantially worse than ResNeXt in ImageNet.

\subsection{Object Detection} We next studied throttleable NNs for the PASCAL VOC 2007 object detection task \citep{everingham2007pascal}. To create a throttleable object detector, we began with the Faster RCNN framework \citep{ren2015faster} and replaced the ``backbone'' CNN with a throttleable network. Our implementation of Faster RCNN is based on the open-source code of \citet{chen2017simple}. We used the DenseNet-169 and ResNeXt-50 models in this experiment. Following the approach of \citet[][Appendix A]{he2016deep} for combining ResNet with Faster RCNN, we split our models after the layer with a $16\times16$ pixel receptive field, using the first half of the network as the feature representation, and the second half as the classifier. The \emph{na\"ive} models are trained on Imagenet and then fine-tuned on VOC2007, with no gating during training. The throttleable models take the throttleable networks from the ImageNet experiments and fine-tune them on VOC2007 with gating.

\paragraph{Results} Similar to results on image classification, we observe that the baseline method achieves higher peak MAP, but its performance quickly drops when any gating is applied. The throttleable models have lower peak MAP, but degrade more gracefully. Interestingly, performance of the throttleable models peaks around $u = 0.5\text{--}0.75$ and then degrades as $u \rightarrow 1$. We also observed this to a lesser extent in some of the classification experiments, most notably with VGG in CIFAR10. It may be that the modules that are only active when $u > 0.75$ receive inadequate training because they are active less often. The result might be improved by altering the distribution of $u$ during training to sample values closer to $1$ more frequently.




\section{Summary and Future Work}
We introduced the idea of a \emph{runtime-throttleable} neural network, which is a single model whose performance can be varied dynamically to produce a range of trade-offs between task performance and resource consumption. We instantiated throttleable NNs using \emph{gated} networks composed of many smaller components that can be turned on and off. In experiments on image classification and object detection, we applied our generic approach to NN throttling to multiple CNN architectures, and found that the resulting models could be effectively throttled across a range of set points, while having peak accuracy comparable to their standard un-throttleable versions. Our proposed \emph{nested gating} scheme was especially effective.

We have three primary objectives for future work. First, we intend to explore more-sophisticated \emph{contextual} gating controllers that take the input into account when making gating decisions \citep[e.g.][]{wu2018blockdrop}. Second, we are investigating using reinforcement learning to train ``meta-controllers'' that manipulate the control signal $u$ directly. This decouples the decision of \emph{how much} to throttle from the decision of \emph{which components} to turn off. Since the meta-controller's learning problem is greatly simplified, it may be possible to train the controller \emph{after deployment} on an edge device. Our goal is to demonstrate adaptive throttling in response to environmental conditions (such as battery charge, illumination level, etc.) in a deployed application. Finally, we are investigating how best to implement the gating mechanism to realize energy savings on real hardware. 

\nottoggle{anonymize}{%
	\section*{Acknowledgements}
	This material is based upon work supported by the Lifelong Learning Machines (L2M) program of the Defense Advanced Research Projects Agency (DARPA) under contract HR0011-18-C-0051, and by the Office of Naval Research (ONR) under contract N00014-17-C-1011. Any opinions, findings and conclusions or recommendations expressed in this material are those of the author(s) and do not necessarily reflect the views of DARPA, the Office of Naval Research, the Department of Defense or the U.S. Government. Special thanks to Dr. Hava Siegelmann, DARPA L2M Program Manager, and Mr. Martin Kruger, ONR Program Manager, for their support and guidance in these programs.
	
	I thank Sek Chai and Aswin Raghavan for many technical discussions and for helping to define the direction of this work.
}

\bibliography{main}
\bibliographystyle{icml2018}

\iftoggle{arxiv}{%
	\appendix
	\section{Experimental Methods}
All of our experiments are implemented using the PyTorch library \citep{paszke2017automatic}, versions 0.3.1, 0.4, and 1.0, and run on Nvidia GTX 1080 Ti and RTX 2080 Ti GPUs.

We generate all of the performance curves by evaluating each model on the full test set using fixed values of $u \in \{0, \frac{1}{16}, \frac{2}{16}, \ldots, 1\}$. The horizontal axes show the \emph{actual} utilization, that is the average proportion of active neurons over the whole test set, which is not always equal to $u$. Each marker on the curves corresponds to a different setting of $u$. We refer to baseline models that do not have gating applied during training as \emph{na\"ive} models.

Each data point in each chart is the result of a single evaluation run for a single instance of the trained model. Therefore there are no error bars shown in the figures. Note that we make no claims that rely on small difference between algorithms. Our conclusions are mainly qualitative, and we believe self-evident from inspecting the data.

\subsection{CIFAR10}
The CIFAR10 dataset \citep{krizhevsky2009learning} is a standard image classification benchmark consisting of 32x32 pixel color images of 10 categories of object. We used the standard 50k image training set and 10k image test set, with no data augmentation. The CNN architectures were as follows. \textbf{DenseNet:} DenseNet-BC with 3 dense blocks having 16 components each with a growth rate of 12. \textbf{ResNeXt:} The ResNeXt architecture for CIFAR as described by \citet{xie2017aggregated} with 16 gated components in each of the 3 stages. \textbf{VGG:} The VGG-D architecture truncated to the first 3 convolution stages followed by a 4096 unit fully-connected layer; all three convolution stages and the fully-connected layer were partitioned into 16 gated components. The ``Independent+Learner'' methods use a ``blind'' control network (FC(1, 32) $\rightarrow$ ReLU $\rightarrow$ FC(32, $n$)) that maps the control input $u$ to gate vectors $\vg$ for each gated module, where $n$ is the number of gated components in the overall network.

In training Phase~1, all models were trained with stochastic gradient descent (SGD) with momentum $0.9$ and weight decay $\SI{5e-4}{}$ for 310 epochs. Batch sizes varied depending on the hardware in use, but were between 64 and 256. We evaluated both the stepwise learning rate schedule used by \citet{xie2017aggregated} and the cosine annealing scheme of \citet{loshchilov2016sgdr} and found that cosine annealing gave consistently better results. The parameters for learning rate annealing were $\eta_{\max} = 0.05$, $\eta_{\min} = 0$, $T_0 = 10$, $T_{\text{mult}} = 2$.

\iftoggle{arxiv}{%
	\newcommand\EqDistRef{\ref{eq:dist}}
	\newcommand\EqTnnObjectiveRef{\ref{eq:tnn-objective}}
	\newcommand\FigCifarBlindRef{Figure~\ref{fig:cifar10-blind}}
}{
	\newcommand\EqDistRef{\ref{main-eq:dist}}
	\newcommand\EqTnnObjectiveRef{\ref{main-eq:tnn-objective}}
	\newcommand\FigCifarBlindRef{Main Figure~\ref{main-fig:cifar10-blind}}
}

In Phase~2, the REINFORCE method was trained with the Adam optimizer \citep{kingma2015adam} with a learning rate of $\SI{1e-3}{}$ for 150 epochs. We used the modified sigmoid activation $\sigma'(x) = \alpha \sigma(x) + (1 - \alpha)(1 - \sigma(x))$ suggested by \citet{wu2018blockdrop} as the output layer for the gating controller to avoid saturation early in training, and annealed $\alpha$ from $0.8$ to $0.99$ over the first 20 epochs. The Concrete method was trained with standard SGD with a fixed learning rate of $\SI{1e-3}{}$ with a temperature parameter of $0.1$, also for 150 epochs. We experimented with annealing the temperature but did not notice any difference. We show results of training using the $C_{\text{dist}}^2$ complexity penalty (\EqDistRef) with $\lambda = 10$ (\EqTnnObjectiveRef).

The performance measure is mean Top-1 classification accuracy on the test set.

\paragraph{Results} The most noticeable result is that \emph{nested gating} substantially outperformed all variations on the independent method for all 3 architectures (\FigCifarBlindRef). The difference is especially pronounced for VGG; we attribute this to VGG learning more ``entangled'' representations than architectures with skip connections, which could make it more sensitive to exactly which components are gated off. For independent gating, the learned gating controllers were consistently better than random gating for both REINFORCE and Concrete training methods. The learned controllers achieve better performance by allocating computation non-uniformly across the different stages of the network (Figure~\ref{fig:usage-cifar10-densenet}). Note that the learned gating functions do not cover the entire range of possible utilization $[0,1]$. The useful range of $u$ was larger for larger $\lambda$ and for complexity penalties with $p = 1$, but these also resulted in lower accuracy overall.

\subsection{ImageNet} 
Our second set of experiments examines image classification on the larger-scale ImageNet dataset \citep{russakovsky2015imagenet} using the DenseNet-169-BC, ResNeXt-50, and ResNet-50 architectures. For ImageNet, we used pretrained weights to initialize the data path, and fine-tuned the weights with gating for 30 epochs using the Adam optimizer \citep{kingma2015adam} with a fixed learning rate of $\SI{1e-5}{}$ with a batch size of 256. We used the DenseNet-169 and ResNet-50 models from the \texttt{torchvision} package of PyTorch, and for ResNeXt-50 we converted the original Torch model \citep{xie2017aggregated} to PyTorch using a conversion utility \citep{clcarwin2017convert}. During fine-tuning, the control parameter $u$ is sampled from $u \sim \text{Uniform}[t, 1]$, where $t$ begins at $1$ and is reduced by $0.05$ every epoch until reaching $0$. We compared this fine-tuning approach to training the gated network from scratch. For both architectures, training from scratch produced substantially worse results than fine-tuning a pretrained model, and we report results only for the fine-tuned models.

We used the standard \texttt{train} / \texttt{val} split from the ILSVRC CLS-LOC dataset. We use the standard proprocessing employed for the PyTorch models, namely normalizing the three image channels to $\mu = (0.485, 0.456, 0.406)$ and $\sigma = (0.229, 0.224, 0.225)$. For training, we augment the data by cropping to a random $224\times224$ image and flipping it horizontally with probability $0.5$. For testing, we resize the image to $256\times256$ and then center-crop to $224\times224$.

The performance measure is mean Top-1 classification accuracy on the \texttt{val} set.

\paragraph{Results}
The throttleable models reached a peak accuracy within about $2-3\%$ of the corresponding pre-trained model, and both were smoothly throttleable through the full range of utilization whereas the pre-trained models degrade rapidly with increased throttling. Unlike in CIFAR10, here the ResNeXt model loses less performance for moderate amounts of throttling compared to DenseNet. Training DenseNet from scratch with gating was less successful, reaching a peak accuracy of only about $50\%$, compared to around $74\%$ for the fine-tuned approach. Because Imagenet is much larger than CIFAR10, we used substantially fewer training epochs and did not use the Cosine Annealing learning rate schedule. It is possible that these two factors were important for achieving good performance when training from scratch on CIFAR10. We did not attempt to train the other models from scratch due to the poor performance of DenseNet.

\begin{figure}[t]
\centering
\includegraphics[width=0.5\columnwidth]{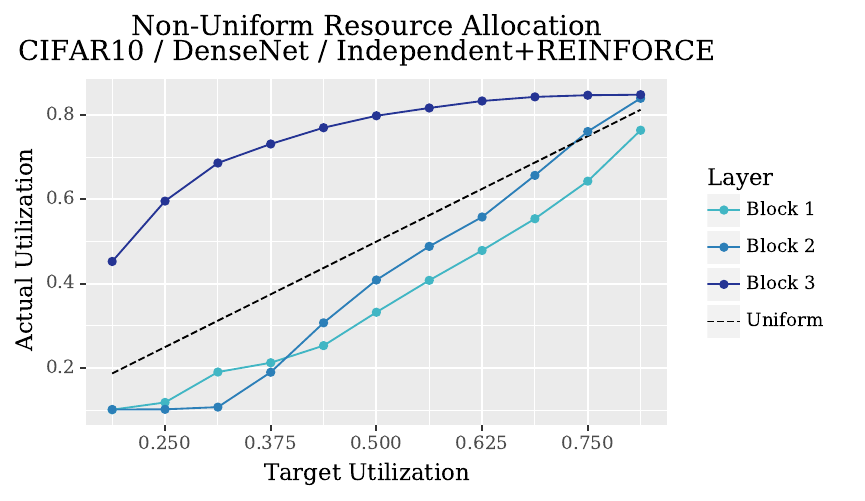}
\caption{The learned pattern of component utilization for DenseNet in CIFAR10 with the REINFORCE training method. Components in the later stages (higher-numbered blocks) are used preferentially over components in earlier stages. The dotted line shows uniform utilization.}
\label{fig:usage-cifar10-densenet}
\end{figure}

\subsection{Object Detection} We next studied throttleable NNs for the PASCAL VOC 2007 object detection task \citep{everingham2007pascal}. To create a throttleable object detector, we began with the Faster RCNN framework \citep{ren2015faster} and replaced the ``backbone'' CNN with a throttleable network. Our implementation of Faster RCNN is based on the open-source code of \citet{chen2017simple}. We used the DenseNet-169 and ResNeXt-50 models in this experiment. Following the approach of \citet[][Appendix A]{he2016deep} for combining ResNet with Faster RCNN, we split our models after the layer with a $16\times16$ pixel receptive field, using the first half of the network as the feature representation, and the second half as the classifier.

We froze the weights from the input layer through the second ``stage'' in the feature networks as well as all batch norm layers \citep{he2016deep}, and fine-tuned the remaining weights on the VOC 2007 \texttt{trainval} images. The \emph{na\"ive} models are trained on Imagenet and then fine-tuned on VOC2007, with no gating during training. The throttleable methods start with the corresponding throttleable architectures trained in the ImageNnet experiment and fine-tunes it on VOC2007. Similar to the Imagenet experiments, during fine-tuning the control parameter $u$ is sampled from $u \sim \text{Uniform}[t, 1]$, where $t$ begins at $1$ and is reduced by $0.1$ every epoch until reaching $0$. Fine-tuning lasts for $16$ epochs for all models.

The implementation of Faster RCNN we used as a base \citet{chen2017simple} uses the parameters described in the original Faster RCNN paper \citep{ren2015faster}, and we did not attempt to optimize them. The performance measure is Mean Average Precision on the VOC2007 \texttt{test} set. We use the mAP implementation from \citet{chen2017simple}, which is derived from the \textbf{VOC 2012} version of the mAP criterion (which differs from the version used for VOC 2007). 

\paragraph{Results} Similar to results on image classification, we observe that the baseline method achieves higher peak MAP, but its performance quickly drops when any gating is applied. The \textsc{Nested} method has lower peak MAP, but degrades more gracefully. Interestingly, performance of the gated variant peaks at $u = 0.75$ and then degrades as $u \rightarrow 1$. We also observed this to a lesser extent in some of the classification experiments, most notably with the VGG architecture in CIFAR10. It is possible that this is due to the modules that are only active when $u > 0.75$ receiving inadequate training, despite the annealing schedule we employed for $u$. The result might be improved by altering the distribution of $u$ during training to sample values closer to $1$ more frequently, and we intend to investigate this.
}

\end{document}